# A Study of Associative Evidential Reasoning*

Yizong Cheng
Department of Computer Science
University of Cincinnati
Cincinnati, OH 45221

Rangasami L. Kashyap
School of Electrical Engineering
Purdue University
West Lafayette, IN 47906

Abstract   Evidential reasoning is cast as the problem of simplifying the evidence-hypothesis relation and constructing combination formulas that possess certain testable properties. Important classes of evidence as identifiers, annihilators, and idempotents and their roles in determining binary operations on intervals of reals are discussed. The appropriate way of constructing formulas for combining evidence and their limitations, for instance, in robustness, are presented.

## 1. Introduction

In this paper, evidential reasoning refers to schemes that relate evidences and hypotheses. More precisely, given an evaluation of certain evidences, an evidential reasoning scheme generates an evaluation of certain hypotheses. When the evaluation of the evidences is a binary one, that is, we either have an evidence or do not have that evidence, the scheme acts as a set function for each hypothesis: a value as an evaluation of the hypothesis is assigned to each subset of evidences.

When the evaluation of hypotheses is also a binary one, the scheme can be represented by a collection of boolean "if-then" rules. Various approaches may be used to make this collection more compact. Intermediate concepts, default rules, and other inventions like the "choice components" in SEEK2 are among these approaches.

The problem becomes more complicated when the evaluation of hypotheses uses values from a linearly ordered set (integers, real numbers, or linguistic quantifiers) or a partially ordered set (intervals or property hierarchies). It becomes even more complex when hypotheses are related to each other (Shafer's theory is an example when hypotheses are subsets of a set), or when the evaluation of evidences are not binary (systems where hypotheses can serve as evidences to other hypothese are examples).

In general, when n evidences are to be dealt by the scheme and all the combinations of these evidences are possible, a complete specification of the scheme for each hypothesis involves $2^n$ values. It is a common practice in many expert systems that instead of having a huge table listing evaluation for each of these combinations, a <u>combination formula</u> is used. Sometimes a combination formula is necessary simply because no complete knowledge about all the combinations is available and the system is built to face all these possibilities. When these formulas are proposed, certain assumptions are made explicitly or implicitly. These assumptions are characterized in the following paragraphs along with the steps of simplification and their consequences on the properties of the formulas.

---------
*Partially supported by the Office of Naval Research under the grant N00014-85K-0611 and NSF under the grant IST 8405052.

The first step in simplifying a set function B(.) is to relate the value of the function on an arbitrary subset to the values on singletons:

$$B(\{e_1,\ldots,e_m\}) = F_m(B(\{e_1\}),\ldots,B(\{e_m\})).$$

Since B is a set function, $F_m$ must obey exchangeability:

$$F_m(\ldots a_k \ldots a_p \ldots) = F_m(\ldots a_p \ldots a_k \ldots).$$

When m=2, this is commutativity.

A common assumption applies when the range of the evaluation of hypotheses is partially ordered. It says that the replacement in a set of evidences S of an evidence by another which favors the hypothesis more will always increase the overall evaluation of the hypothesis. In terms of the function $F_m$, this is the monotonicity property:

$$F_m(\ldots a \ldots) \leq F_m(\ldots b \ldots) \quad \text{if } a \leq b.$$

Under the "choice component" interpretation, SEEK is not a monotonic system, since replacing a "minor finding" by a "major finding" does not necessarily increase the diagnosis level of a disease.

After this first step, a specification of $2^n$ terms for each hypothesis is reduced to a specification of 2n terms (n $F_m$ formulas and n $F_1$ values). This is further simplified when the second step is taking place:

$$F_m(a_1,\ldots,a_m) = F_2(F_{m-1}(a_1,\ldots,a_{m-1}),a_m).$$

We will write $F_2$ in the form of a binary operator *:

$$a*b = F_2(a,b).$$

After these two steps, we know that the operator * must be associative:

$$(a*b)*c = a*(b*c).$$

A major merit of associativity is that as evidences accumulate, we can continuously update the evaluation using only the latest evidence and the current evaluation. The original version of the certainty factor updating rule in MYCIN is not associative.

When * is a binary operator on an ordered set, monotonicity and associativity make the set an ordered semigroup under *. When the ordered set is an interval of real numbers, and the operator is a continuous one, this semigroup is an example of connected ordered topological semigroups (Clifford, 1958). The continuity of * may be stated as: if $a*b \leq d \leq a*c$, then there exists x, $b \leq x \leq c$, such that $a*x = d$.

There are two approaches to design a combination formula for a system. The first approach starts with the nature of the evaluation of hypotheses, and the formula is constructed through a theory for the evaluation scheme and a set of assumptions. One example is the interpretation of the evaluation as probability assignments and derivation of formulas based on well-accepted theories of probabilities. The second approach starts with required properties of the combination and the formula is constructed based on their properties. This approach is akin to solving a functional equation in measurement theory, or constructing a model to mimic the enviroment in machine learning. We hope this paper will provide insights to this second approach.

## 2. Special Classes of Evidences

From now on, we assume the above simplification steps and restrict ourselves to commutative, associative, monotonic, and continuous binary operations defined on an interval of real numbers. For the evaluation of a hypothesis, those evidences $e_m$ that produce the same $B(\{e_m\})$ are indistinguishable and interchangeable, and they form an equivalence class.

There are two important classes of evidences namely the identities and the annihilators. Choosing them is the first step one should go in designing a combination formula. An evidence is an identity when its state of presence is irrelevant to the hypothesis. The evaluation produced by this evidence to the hypothesis should be a real number e that makes $a*e=a$ for all possible evaluation values a. e is unique by commutativity and associativity. It divides the range of evaluation into the positive side and the negative side. From monotonicity, evidences on either side are reinforcing each other. That is, when $a,b \geq e$, $a*b \geq \max\{a,b\}$; when $a,b \leq e$, $a*b \leq \min\{a,b\}$.

An evidence is an annihilator if its presence settles down the evaluation of the hypothesis once and forever. In terms of a real number z, this means $a*z=z$ for all a. If one considers only the positive side or the negative side, the endpoint of either side is the annihilator of that side.

Identities and annihilators are special cases of idempotents. If $u*u=u$, then u is an idempotent. By monotonicity, idempotents divide the range into basic segments closed under the binary operation. Idempotents also determines the binary operation of

264

real points separated by them, as long as these points are both on the positive side or both on the negative side. Consider u as an idempotent in the positive side [e,z]. Let e≤a≤u≤b≤z. Since u*u =u≤b≤u*z=z, by continuity, there exists a positive x such that u*x=b. Thus, a*b≤u*b=u*u*x=u*x=b. Because this is the positive side, a*b≥b. Hence, a*b=b=max{a,b}. Similarly, the only operation on negative points separated by an idempotent is the min operation. Particularly, when all the points are idempotents, that is, the combination is closed on equivalence classes of evidences, the only combination formula is max on the positive side and min on the negative side.

## 3. Combination Rules for Basic Segments

We do not have much choice for the binary operation inside segments delimited by idempotents either. Each segment on the positive side has its lower endpoint as its identity and its upper endpoint as its annihilator and they are the only idempotents in the segment. Let <S1,*1,<1> and <S2,*2,<2> be two ordered semigroups. They are o-isomorphic to each other if there exists a bijection h: S1->S2 such that for any a,b in S1, h(a) *2 h(b) = h(a *1 b) and h(a) <2 h(b) if a <1 b. Faucett (1955) proved that a positive basic segment is o_isomorphic to [0,oo] under addition provided there are no interior nilpotents (points repeatedly combined with themselves becoming the annihilator). Mostert and Shields (1957) proved that otherwise it is o-isomorphic to [0,1] under the bounded sum, a*b = min{ a+b, 1 }. A negative basic segment is merely the dual of a positive one. Once h is given, the binary operation on S1 is given as

$$a *1 b = h^{-1}(h(a) *2 h(b)).$$

When *2 is addition, $a*b = h^{-1}(h(a)+h(b))$. When *2 is the bounded sum, $a*b = h^{-1}(\min\{h(a)+h(b),1\})$.

Let S1=[0,1] and S2=[0,oo]. A particularly interesting family of bijections h is the Hamacher operators (Zimmermann, 1978):

$$h(a) = \log\left(\frac{r}{1-a} + 1 - r\right), \quad r>0.$$

The corresponding *1 operations are

$$a * b = \frac{a + b + (r-2)ab}{1 - (r-1)ab}, \quad r>0.$$

a*b is an increasing function of r. Some r values lead to simple and familiar rational formulas. For example, when r=1, the Hamacher operator is

$$a * b = a + b - ab,$$

265

which was proposed by James Bernoulli in 1713 for combining evidence (Shafer, 1978). The Hamacher operator with r=2 is the relativistic velocity addition formula when the speed of light is normalized to unity:

$$a * b = \frac{a + b}{1 + ab}.$$

Because of the r values, the combination result from the velocity addition is always greater than that from Bernoulli's rule.

A family of o-isomorphisms is $h(a) = a^p$, p>0. It generates a family of bounded-sum type operations:

$$a * b = \min \{ 1, (a^p + b^p)^{1/p} \}.$$

## 4. The Concatenation of Symmetric Positive and Negative Segments

By now, we have studied the combination rule within a basic segment and among segments on the positive or the negative side. In this section, we study the case when a positive segment is concatenated with its negative dual. In other words, we study the cross combination of a positive evidence and a negative evidence.

Two dual basic segments can joint together at there common identity point or their common annihilator pointor. For the latter case, the cross combination becomes extremely simple: By monotonicity, the only cross combination is a*b=z, where z is the common annihilator. When all the points are idempotents, from this result and the reinforcing rule, the only combination rule is a*b = median (a,b,z). This result has been mentioned by Silvert(1979) in connection with stable combinations.

The cross combination rule is often quite different from the rule on either side. Consider the concatenation of [0,1/2] with [1/2, 1] with 1/2 the common annihilator. Suppose

$$a * b = \frac{4ab - 1}{4(a + b - 1)}$$

is the combination on [1/2,1], then its dual form is exactly the same on [0,1/2]. (This is isomorphic to the harmonic average, and no identity exists on either side.) However, the only possible cross combination is, as shown above, a*b=1/2.

Let us consider the other situation, when the joint is the common identity e. Let f be the dual mapping from the negative segment onto the positive segment, that is, when $a,b \leq e$, $f(a*b)= f(a)*f(b)$. The following theorem says that when the segment combination is the addition type, the cross combination is also determined, provided that $a*f(a)=e$ for all $a$ except the endpoints.

Theorem. Let h be the o-isomorphism from the positive basic segment onto the addition semigroup of $[0,\infty]$. Let $a*f(a)=e$ for all a except the endpoints. Then the combination rule for the complete interval is

$$a * b = g^{-1}(g(a) + g(b))$$

where

$$g(a) = h(a) \quad \text{if } a \geq e$$
$$\phantom{g(a) =} -h(f(a)) \quad \text{if } a < e.$$

The combination of the two endpoints is left undefined.

The proof of this theorem is omitted due to lack of space.

Consider the range $[-1,+1]$. Suppose h is the Hamacher operator and $f(a)=-a$. The combination in $[-1,1]$ is

$$a * b = \frac{a + b + (r-2)ab}{1 + (r-1)ab}, \quad a,b \geq 0$$

$$\frac{a + b - (r-2)ab}{1 + (r-1)ab}, \quad a,b \leq 0$$

$$\frac{a + b}{1 + a + (1-r)a(1-b)}, \quad a<0<b \text{ and } -a \leq b$$

$$\frac{a + b}{1 - b - (1-r)b(1+a)}, \quad a<0<b \text{ and } -a > b$$

When $r=2$, this collapses into one formula:

$$a * b = \frac{a + b}{1 + ab}.$$

Thus, the velocity combination applies to both directions. When $r=1$, the combination is an associative extension of Bernoulli's rule:

$$a * b = a + b - ab \quad a,b \geq 0$$

$$\phantom{a * b =} a + b + ab \quad a,b \leq 0$$

$$\phantom{a * b =} (a+b)/(1+b) \quad a<0<b \text{ and } -a \leq b$$

$$\phantom{a * b =} (a+b)/(1-b) \quad a<0<b \text{ and } -a > b.$$

The original version of MYCIN's combination rule for certainty factors is not associative. The above is the only way to extend it associatively to the whole range.

## 5. Concatenation of Non-Symmetric Positive and Negative Segments

It may not be always desirable to have the midpoint of a range as the identity and the positive side and the negative side with the same length. For example, when probabilities in the range of [0,1] are used, if the overall probability of a disease is 0.001, then a finding that by itself gives a probability of 1/2 is definitely a very positive support for the diagnosis of the disease. In this section, we investigate combinations with arbitrary points as identities.

Let us consider the range of [0,1]. Let g be the isomorphism generated from the Theorem that requires 1/2 to be the identity. Let s be the desired identity (0.001 in the above example). Suppose y is an order-preserving mapping from [0,1] onto [0,1] that maps s to 1/2. Then g.y is the bijection from [0,1] to the real line that generates an addition-type combination with s as the identity. To find the simplest rational mapping y, let $y(a) = (A+Ba)/(C+Da)$, with A, B, C, and D being constants to be determined. From $y(0)=0$, we must have $A=0$. From $y(1)=1$, we have $B=C+D$. $y(s)=1/2$ requires that $2sB=C+Ds$. This gives the relation $B/D = (1-s)/(1-2s)$ and

$$y(a) = \frac{(1-s)a}{s + (1-2s)a}.$$

Suppose $g(a)=\log(1/a -1)$, or the Hamacher operator from [0,1] to the real line when $r=2$. Then the composed isomorphism is

$$g.y(a) = \log \frac{(1-a)s}{a(1-s)}$$

and the combination rule thus generated is

$$a * b = \frac{ab}{ab + (1-a)(1-b)\,s/(1-s)}.$$

This is the formula for calculating conditional probability $P(h|e_1 \ldots e_m)$ from $P(h|e_1),\ldots,P(h|e_m)$ when the prior is $P(h)=s$ and the conditional independence is assumed. When $s=1/2$, we go back to the symmetric case which is actually the [0,1] version of relativistic velocity addition.

## 6. About Robustness

The maximum slope of the Hamacher family is 1 when $r \leq 2$ and $r^2/4(r-1)$ when $r>2$. Unless a very large r is chosen, the slope is

268

well-bounded and the operation is considered robust.

On the other hand, cross combinations generated through the Theorem are never robust. Let a be a value close to the lower endpoint and b close to f(a), say, b=f(a)+d. Then

$$a * b = h^{-1}( h(b)-h(f(a)) ).$$

Since both h(b) and h(f(a)) are very large values, the difference can be any possitive value. In other words, a*b can be any value in the entire positive segment [e,z]. A small change in b from f(a) to f(a)+d can change a*b substantially.

Although the cross combination for certainty factors in the original MYCIN is not associative, it is robust. The only way to achieve robustness and associativity for cross combination is to use bounded-sum type operations.

## 7. Conclusion

This paper shows a possible approach of developing combination formulas for evidential reasoning systems. Results on ordered semigroups are used to demonstrate all the possibilities one can have. A theorem on jointing segments together to form more interesting evaluation ranges is developed. Some important examples as variations of the Hamacher operator family are shown. Robustness of these rules is studied.

Further work along this approach with different types of evaluation and reasoning structure ( exemplified in the Introduction) is needed. Connection with other areas, for example, machine learning, should be explored.